\pdfoutput=1
\documentclass[11pt]{article}

\usepackage[]{ACL2023}

\usepackage{times}
\usepackage{latexsym}

\usepackage[T1]{fontenc}
\usepackage[utf8]{inputenc}

\usepackage{microtype}

\usepackage{inconsolata}
\usepackage{multirow}
\usepackage{graphicx}
\usepackage{adjustbox}
\usepackage{amsmath}
\usepackage{amssymb}
\usepackage[cjk]{kotex}
\usepackage{booktabs}
\usepackage{comment}
\usepackage{makecell}
\usepackage{xcolor}

\usepackage{floatrow}
\usepackage{graphicx}
\usepackage[label font=bf]{subfig}
\usepackage{caption}
\usepackage{fancyvrb}
\usepackage{fvextra}

\usepackage{adjustbox}
\usepackage{graphicx}
\usepackage{arydshln}
\definecolor{cusorange}{RGB}{255, 50, 50}
\definecolor{medblue}{RGB}{28, 98, 215}
\definecolor{darkgreen}{RGB}{0, 100, 0}
\definecolor{darkred}{RGB}{167, 51, 51}

\title{Evaluation of Question Generation Needs More References}

\author{
\quad Shinhyeok Oh\thanks{\quad Equal Contribution.}
\quad Hyojun Go\footnotemark[1]
\quad \textbf{Yunsung Lee} 
\quad \textbf{Hyeongdon Moon} \\
\quad \textbf{Myeongho Jeong} 
\quad \textbf{Hyun Seung Lee} 
\quad \textbf{Seungtaek Choi}\thanks{\quad Corresponding author.} \\
Riiid AI Research\\
\texttt{\{shinhyeok.oh, hyojun.go, seungtaek.choi\}@riiid.co}, \\
}

\begin{document}

\newcommand{\todoc}[2]{{\textcolor{#1}{\textbf{#2}}}}
\newcommand{\todoblue}[1]{\todoc{blue}{\textbf{#1}}}
\newcommand{\todored}[1]{\todoc{red}{#1}}
\newcommand{\todoorange}[1]{\todoc{orange}{#1}}
\newcommand{\todopurple}[1]{\todoc{purple}{#1}}

\newcommand{\hist}[1]{\todored{\textbf{hist:} #1}}
\newcommand{\spilly}[1]{\todoorange{spilly: #1}}
\newcommand{\shinhyeok}[1]{\todoblue{shinhyeok: #1}}
\newcommand{\todogreen}[1]{\todoc{green}{#1}}
\newcommand{\hyojun}[1]{\todogreen{\textbf{hyojun:} #1}}
\newcommand{\yunsung}[1]{\todopurple{\textbf{yunsung:} #1}}

\newcommand\Tstrut{\rule{0pt}{2.2ex}}       % "top" strut
\newcommand\Bstrut{\rule[-0.6ex]{0pt}{0pt}} % "bottom" strut
\newcommand{\TBstrut}{\Tstrut\Bstrut} % top&bottom struts

\maketitle
\begin{abstract}
\label{sec:abstract}

Question generation (QG) is the task of generating a valid and fluent question based on a given context and the target answer. According to various purposes, even given the same context, instructors can ask questions about different concepts, and even the same concept can be written in different ways. However, the evaluation for QG usually depends on single reference-based similarity metrics, such as n-gram-based metric or learned metric, which is not sufficient to fully evaluate the potential of QG methods. To this end, we propose to paraphrase the reference question for a more robust QG evaluation. Using large language models such as GPT-3, we created semantically and syntactically diverse questions, then adopt the simple aggregation of the popular evaluation metrics as the final scores. Through our experiments, we found that using multiple (pseudo) references is more effective for QG evaluation while showing a higher correlation with human evaluations than evaluation with a single reference.

\end{abstract}
\section{Introduction}
\label{sec:introduction}

Question generation (QG) is the task of generating questions that are relevant to and answerable by given text. 
Since QG can be applied in not only educational scenarios~\cite{kurdi2020systematic, steuer2021not, moon-etal-2022-evaluating} but also improving question-answering tasks~\cite{chen2021factuality,wang2018learning, yu2020based}, designing better QG frameworks and their automatic evaluations have gained more attention~\cite{https://doi.org/10.48550/arxiv.2210.11536, https://doi.org/10.48550/arxiv.2210.03992}.

However, previous QG works mostly evaluate their methods based on how similar the generated questions are to the gold reference questions~\cite{chan-fan-2019-recurrent, DBLP:conf/nlpcc/ZhouYWTBZ17, du-cardie-2018-harvesting}, using n-gram-based similarity metrics, such as BLEU~\cite{papineni-etal-2002-bleu} and ROUGE~\cite{lin-2004-rouge}. 
Given a single reference, these metrics do not account for the lexical and semantic diversity of questions~\cite{DBLP:conf/iclr/ZhangKWWA20}, showing poor correlation with human judgment~\cite{liu-etal-2016-evaluate, novikova-etal-2017-need, chaganty-etal-2018-price}.
Though prior works studied alternative metrics of leveraging language models, such as BERTScore~\cite{DBLP:conf/iclr/ZhangKWWA20} and BLEURT~\cite{sellam-etal-2020-bleurt}, such metrics are limited in that the diversity of gold questions is only implicitly represented in the embedding space, rather than data space (or, raw questions).

To explicitly compare with the diverse gold questions in the data space, we propose to augment the single reference question for evaluating QG frameworks, which we call Multi-Reference Evaluation (MRE), by leveraging the few-shot ability of large language models (LLMs) like GPT-3~\cite{NEURIPS2020_1457c0d6} and ChatGPT~\cite{openai_2023}. 
Though there have been efforts to augment references for improving evaluations, they are either limited in other text generation tasks, such as machine translation~\cite{bawden-etal-2020-study} and question answering~\cite{liu-etal-2021-language}, or the methods are hard to be applied in question generation tasks, as naive LLMs generate some negative (toxic or erroneous) questions~\cite{10.1007/978-3-031-11644-5_13}.
Therefore, we utilize LLMs for paraphrasing to augment a reference question, rather than generating new questions from the given context. 
To the best of our knowledge, we are the first to apply reference augmentation to evaluate the QG frameworks. We briefly summarize our main contributions as follows:

\begin{itemize}
\item We propose to augment the single reference for multiple reference evaluation (MRE) that can explicitly consider syntactic and semantic variations of questions.
Experimental results on quiz design dataset~\cite{laban2022quiz} show that the performance of existing metrics can be considerably improved when MRE is applied.
\item MRE is metric-agnostic, such that various metrics can be improved with our method. Since each existing metric can discover different insights, such as BLEU for lexical similarity and BERTScore for semantic similarity, MRE can improve these multiple various lenses for investigating QG frameworks. 
\item We release the augmented reference questions as supplementary materials\footnote{to be published in Findings of ACL2023}, which provide an opportunity to reproduce our results for further research. We further validated whether the augmented references are correct or not by human annotators. 

\end{itemize}

\section{Methodology}
\label{sec:approach}

\subsection{Single Reference Evaluation (SRE)}
Previous works for QG evaluation measure the quality of a generated question $q^g$ in regards to a gold reference question $q^r$ as $M(q^g, q^r)$, where $M$ denotes a similarity metric that is widely used in QG evaluation such as BLEU and ROUGE-L.
However, since these metrics suppose only one gold reference, even an appropriate question can be assigned a low score, namely \textit{false positive} problem.

\subsection{Multi-Reference Evaluation (MRE)}
To deal with this problem, we propose the multi-reference evaluation, where the candidate question $q^g$ is compared with multiple references $Q=\{q^r_0, q^r_1,\dots, q^r_N\}$:
\begin{equation}
    s = \max_i M(q^r_i, q^g) \quad \text{for} \quad i=0,\dots,N.
\end{equation}

By comparing more diverse gold questions with existing metrics, we can measure the more realistic ability of QG frameworks. Note that, as our method could adopt any similarity-based metrics, we can better gain useful insights from various metrics showing different characteristics of generated questions.

However, as it is impractical to collect such multiple references with human annotators, we leverage the recent large language models, specifically GPT-3 and ChatGPT, such that replace $Q$ with $\hat{Q}$. Given a reference question $q^r_0$, we augment it with $N$ questions:
\begin{equation}
    \hat{Q} = \text{LLM}(q^r_0).
\end{equation}

Note that we give a gold question $q^r_0$ only, rather than the pair of context and question as in~\cite{liu-etal-2021-language}.
It is because the zero-shot QG ability of LLMs is reportedly risky for educational purposes~\cite{10.1007/978-3-031-11644-5_13}.
We thus use LLMs as a paraphrase generator, which reportedly works well since there is a high correlation between paraphrasing and training paradigms about LLM~\cite{chen2022zero}.

As GPT-3 is inferior to ChatGPT in the zero-shot settings, here we employ the in-context learning ability of GPT-3, where we give three ChatGPT-paraphrased questions questions as a demonstration for GPT-3 like Appendix~\ref{appsec:prompts}.
We will further investigate the correctness of the paraphrased questions in experiments (Section~\ref{sec:human}).

\section{Experiments}
\label{sec:experiments}
\begin{table*}[t]\centering
\resizebox{\textwidth}{!}{
\begin{tabular}{lcccccccccccc}
\toprule 
 & & \multicolumn{5}{c}{Pearson Correlation} & & \multicolumn{5}{c}{Spearman Correlation}  \\
 \cline{3-7} \cline{9-13}
 & & \multirow{2}{*}{\begin{tabular}[c]{c}\\SRE\\\end{tabular}} & \multicolumn{4}{c}{MRE} & & \multirow{2}{*}{\begin{tabular}[c]{c}\\SRE\\\end{tabular}} & \multicolumn{4}{c}{MRE} \Tstrut \\
 \cline{4-7} \cline{10-13}
 & & & HRQ-VAE & \begin{tabular}[c]{c}GPT-3\Tstrut\\(0-shot)\end{tabular} & \begin{tabular}[c]{c}GPT-3\Tstrut\\(3-shot)\end{tabular} & \begin{tabular}[c]{c}ChatGPT\Tstrut\\(0-shot)\end{tabular} & & & HRQ-VAE & \begin{tabular}[c]{c}GPT-3\Tstrut\\(0-shot)\end{tabular} & \begin{tabular}[c]{c}GPT-3\Tstrut\\(3-shot)\end{tabular} & \begin{tabular}[c]{c}ChatGPT\Tstrut\\(0-shot)\end{tabular} \\
\midrule
BLEU-4 & & 0.2028 & 0.2443 & 0.2782 & 0.3162 & \textbf{0.3630} & & 0.2772 & 0.3224 & 0.2688 & 0.3021 & \textbf{0.3340}  \\
ROUGE-L & & 0.2908 & 0.3325 & 0.3241 & 0.3447 & \textbf{0.3799} & & 0.2787 & 0.3270 & 0.3050 & 0.3330 & \textbf{0.3637} \\
RQUGE & & 0.2932 & - & - & - & - & & 0.2571 & - & - & - & -\\
METEOR & & 0.3447 & 0.2968 & 0.3480 & 0.3877 & \textbf{0.4116} & & 0.3111 & 0.2822 & 0.3159 & 0.3562 & \textbf{0.3780}\\
BERTScore & & 0.3556 & 0.3634 & 0.3552 & 0.3877 & \textbf{0.4033} & & 0.3462 & 0.3568 & 0.3327 & 0.3723 & \textbf{0.3859}\\
MoverScore & & 0.4383 & 0.3835 & 0.4297 & 0.4693 & \textbf{0.4953} & & 0.3882 & 0.3643 & 0.3885 & 0.4214 & \textbf{0.4292}\\
BLEURT & & \underline{0.4739} & \underline{0.4287} & \underline{0.4656} & \underline{0.4803} & \textbf{\underline{0.5019}} & & \underline{0.4566} & \underline{0.4193} & \underline{0.4456} & \underline{0.4648} & \textbf{\underline{0.4816}} \\
\bottomrule
\end{tabular}
}
\caption{Results of the correlation coefficient between measured metrics and human score. The best scores in methodology are in bold, and the best scores in metrics are underlined. These depend on the types of correlation measures. ‘-’ denotes unreported results.}
\label{table:correlation}
\end{table*}

\subsection{Dataset and Evaluation}
To verify the effectiveness of MRE, we use quiz design dataset~\cite{laban2022quiz} for measuring the correlation between automatic question evaluation and human annotation.
The quiz design dataset includes 3,164 human-annotated samples, which consist of context, answer, and automatically generated questions.
For each sample, the human annotates whether the question is fluent, able to derive the given answer, and fits the context (1) or not (0).

We define the gold human score of a question as the average of the discrete human annotations in $[0, 1]$.
Then, we select questions with a human score of 1 as the reference question for the given passage.
Finally, for the remaining questions, we measure the Pearson correlation coefficient~\cite{freedman2007statistics} and Spearman's rank correlation coefficient~\cite{zar2005spearman} between the human score and automatic evaluation scores.

\subsection{Metrics}
Here, as we aim to enhance the existing QG evaluation metrics with multi-reference evaluation, we choose widely used metrics to apply multi-reference evaluation.
We apply multi-reference evaluation to BLEU-4~\cite{papineni-etal-2002-bleu}, ROUGE-L~\cite{lin-2004-rouge}, METEOR~\cite{banerjee-lavie-2005-meteor}, BERTScore~\cite{DBLP:conf/iclr/ZhangKWWA20}, BLEURT~\cite{sellam-etal-2020-bleurt}.
Also, we add RQUGE~\cite{mohammadshahi2022rquge}, which is a reference-free QG evaluation metric, as our baseline.
We briefly summarize the metrics used in our experiments as follows:
\begin{itemize}
\item \textbf{BLEU-4}~\cite{papineni-etal-2002-bleu} is a metric that utilizes n-gram precision to evaluate the similarity between a generated text and a reference text. The metric counts the number of occurrences of unigrams, bigrams, trigrams, and four-grams that match their corresponding counterparts in the reference text.
\item \textbf{ROUGE-L}~\cite{lin-2004-rouge} is a metric that utilizes unigram recall to evaluate the similarity between a generated text and a reference text. The metric counts the length of the longest common subsequence as the numerator rather than the exact number of matches.
\item \textbf{RQUGE}~\cite{mohammadshahi2022rquge} first predicts answer span with question answering model then computes score with scorer module from given generated question, gold answer, and context. Since RQUGE does not depend on a reference question for evaluation, we only report the correlation of the original RQUGE.
\item \textbf{METEOR}~\cite{banerjee-lavie-2005-meteor} measures a score by using a combination of unigram-precision, unigram-recall, and fragmentation measures.
\item \textbf{BERTScore}~\cite{DBLP:conf/iclr/ZhangKWWA20} utilize contextual embeddings for compute token similarity. We report BERTScore based on \texttt{roberta-large}.
\item \textbf{BLEURT}~\cite{sellam-etal-2020-bleurt} 
is a trained metric using a regression model trained on rating data. It combine expressivity and robustness by pre-training a fully learned
metric on large amounts of synthetic data, before
fine-tuning it on human ratings.
\end{itemize}

\begin{table}
\centering
\resizebox{0.75\columnwidth}{!}{
\begin{tabular}{c!{\vrule width \lightrulewidth}cc} 
\hline
Model                                                     & \begin{tabular}[c]{@{}c@{}}Same\\answer\end{tabular} & \begin{tabular}[c]{@{}c@{}}Same\\meaning\end{tabular}  \\ 
\hline
GPT-3 (0-shot) & 0.77 & 0.79 \\
\hline
GPT-3 (3-shot) & 0.83 & 0.83 \\ 
\hline
ChatGPT (0-shot) & 0.92 & 0.93 \\

\hline
\end{tabular}}
\caption{
Human evaluation results of whether paraphrased question by the LLM has the same correct answer and meaning as the reference question.}
\vspace{-0.2cm}
\label{tab:human-paraphrased}
\end{table}

\subsection{Implementation details}
We implemented the paraphrasing frameworks by using two LLMs: OpenAI GPT-3 API~\cite{NEURIPS2020_1457c0d6} and ChatGPT Webservice~\cite{openai_2023}. For GPT-3, we set the model as "text-davinci-003" and the temperature as 0.5. For ChatGPT, we utilized the default setting since we cannot control it. Our prompts are described in Appendix~\ref{appsec:prompts}. We made 20 examples by using LLMs. For additional comparisons with the fine-tuned paraphrasing model, we also implemented HRQ-VAE~\cite{hosking-etal-2022-hierarchical}.

\begin{table*}[t]\centering
\resizebox{0.8\textwidth}{!}{
\begin{tabular}{ccccccc}
\toprule
& \multicolumn{6}{c}{$\Delta (MRE - SRE)$} \\
\cmidrule{2-7} 
Human Score & BLEU-4 & ROUGE-L & METEOR & BERTScore & MoverScore & BLEURT\\
\midrule
1 & $+$\ 0.2267 & $+$\ 0.1221 & $+$\ 0.1034 & $+$\ 0.0592 & $+$\ 0.0439 & $+$\ 0.0400\\
\midrule
0 & $+$\ 0.0350 & $+$\ 0.0846 & $+$\ 0.0941 & $+$\ 0.0398 & $+$\ 0.0190 & $+$\ 0.0373\\
\bottomrule
\end{tabular}}
\caption{Score changes with multiple reference evaluation $\Delta (MRE - SRE)$ through ChatGPT for questions of human score 0 and 1. 
}
\vspace{-0.2cm}
\label{table:ablation}
\end{table*}

\subsection{Main Results}
As shown in Table~\ref{table:correlation}, we empirically validate the following observations of the advantages of diversified multi-reference evaluation: 1) Our multi-reference evaluation tends to improve the correlation between human score and evaluation metrics. 2) On LLMs, correlation with the human score is high in the order of ChatGPT (0-shot), GPT-3 (3-shot), and GPT-3 (0-shot) paraphrasing framework.
Specifically, GPT-3 (3-shot) and ChatGPT paraphrasing framework considerably improve both Pearson correlation and Spearman correlation for all metrics, while paraphrasing with GPT-3 (0-shot) and HRQ-VAE failed at increasing correlations of some metrics.

Also, the increase in correlation through MRE is related to the performance of the paraphrasing framework.
As shown in Table~\ref{tab:human-paraphrased}, the paraphrase of the reference question is better in the order of ChatGPT, GPT-3 (3-shot), and GPT-3 (0-shot). 
Considering the effect of MRE is also in the same order, we conjecture that the performance of the paraphrasing framework is also important for the effect of MRE.
More details in Table~\ref{tab:human-paraphrased} are described in Section~\ref{sec:human}.

\begin{figure}[t]\centering
\includegraphics[width=0.96\linewidth]{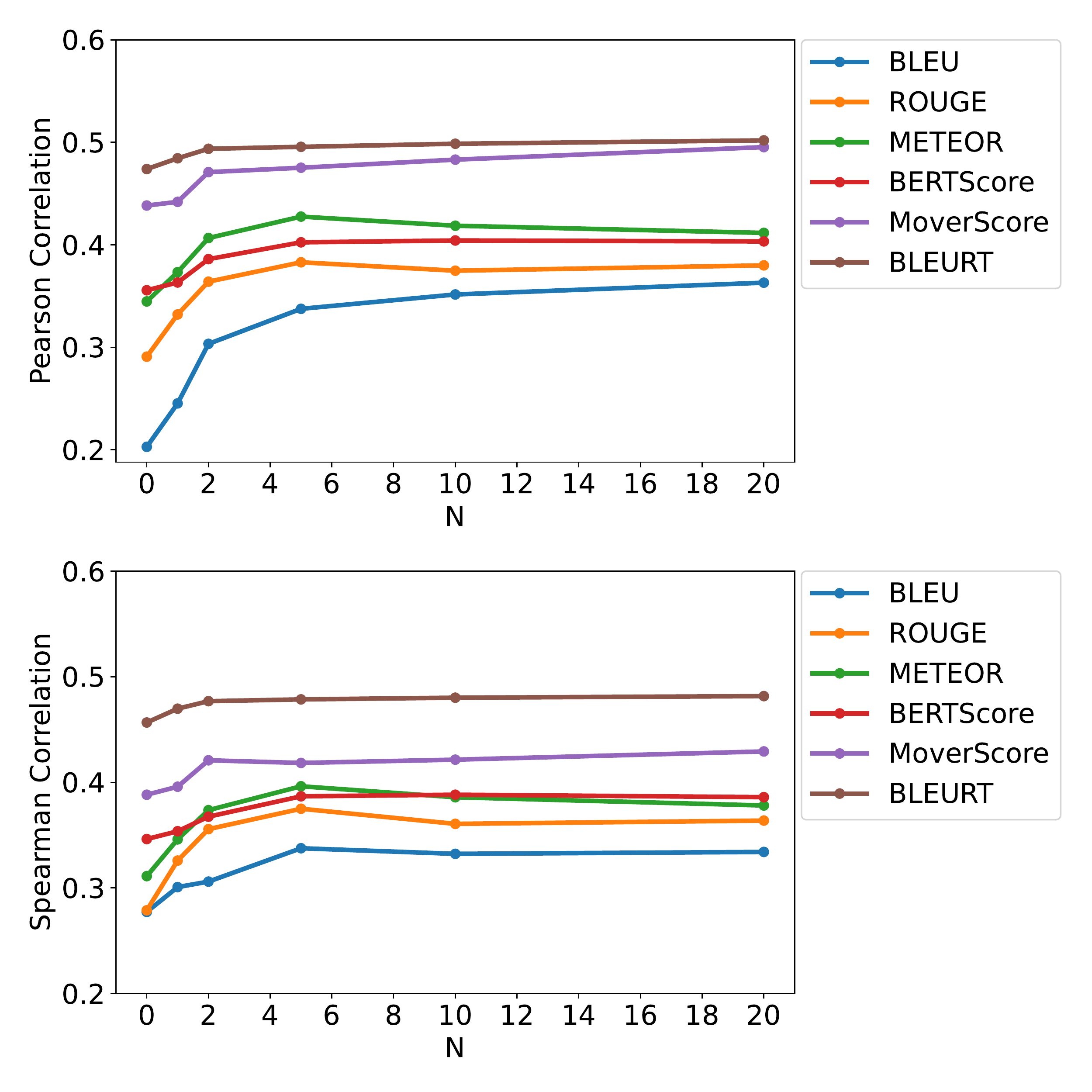}
\caption{Changes of Pearson and Spearman Correlation coefficients on the number of references generated by ChatGPT (0-shot).}
\label{fig:pearson_spearman_by_n}
\vspace{-0.4cm}
\end{figure}

\subsection{Analysis for MRE}
\paragraph{The effect of $N$}
We analyze the effect of the number of reference questions $N$ by changing $N$ to 1, 2, 5, 10, and 20. 
Figure~\ref{fig:pearson_spearman_by_n} shows the change of the correlation coefficient according to the change of $N$.
The results show that even if only one augmented reference question is used, the correlation is higher than that of the single reference evaluation.
Also, if more augmented reference questions are used, the correlation with the human score increases and becomes saturated when $N$ exceeds a certain level ($N \approx 5$).

\paragraph{Score change with multi-reference evaluation}
We further explore how MRE changes original metrics.
Specifically, we report average score differences between the original metric and the multi-reference version of it with ChatGPT for accepted and unaccepted candidate questions.
Questions with the human score of 1 and 0 are considered accepted questions and unaccepted questions, respectively.

As shown in Table~\ref{table:ablation}, multi-reference evaluation increases the score of accepted questions relatively more than that of an unaccepted question.
For example, BLEU-4 score increases by 0.2267 for accepted questions, compared to 0.0350 for unaccepted questions.
These results mean that multi-reference evaluation makes original metrics more correlated with the human score by enlarging the score of acceptable questions than unacceptable questions.

\begin{table*}[t]
\centering
\resizebox{0.98\textwidth}{!}{
\renewcommand{\arraystretch}{1.1}
\begin{tabular}{llllccccc}
\toprule

 & Generated Question & Approach & Reference Question & B-4 & R-L & BS & BR & Human \\
\midrule
\multirow{5}{*}{E1} & \multirow{5}{*}{\begin{tabular}[c]{@{}l@{}} What is the definition of \\sustainable energy?\end{tabular}}
      & SRE  & What does it mean if energy is sustainable?  & 0.00 & 0.27 & 0.68 & 0.75 & \multirow{5}{*}{1.00} \\
      \cdashline{3-8}
   & & MRE-B-4 & What is the definition of sustainable energy?  & 1.00 & - & & \\
      \cdashline{3-8}
     & & MRE-R-L & What is the definition of sustainable energy?  & - & 1.00 & & \\
      \cdashline{3-8}
   & & MRE-BS & What is the definition of sustainable energy?  & - & - & 1.00 & - \\
      \cdashline{3-8}
   & & MRE-BR & What is the definition of sustainable energy?  & - & - & - & 0.97 \\
        
\midrule
 \multirow{5}{*}{E2} & \multirow{5}{*}{\begin{tabular}[c]{@{}l@{}} What are some examples of \\renewable energy sources?\end{tabular}}
      & SRE  & What are some renewable energy sources?  & 0.00 & 0.86 & 0.87 & 0.83 &\multirow{5}{*}{1.00}\\
      \cdashline{3-8}
    & & MRE-B-4 & What are some examples of renewable energy?  & 0.53 & - & &\\
      \cdashline{3-8}
    & & MRE-R-L & What are some examples of alternative energy sources?  & - & 0.87 & &\\
      \cdashline{3-8}
    & & MRE-BS & What are some examples of renewable energy?  & - & - & 0.95 & - \\
      \cdashline{3-8}
    & & MRE-BR & What are some examples of renewable energy?  & - & - & - & 0.85\\
\midrule
 \multirow{5}{*}{E3} & \multirow{5}{*}{\begin{tabular}[c]{@{}l@{}} How is energy sustainable?\end{tabular}}
      & SRE  & What does it mean if energy is sustainable?  & 0.00 & 0.33 & 0.74 & 0.77 &\multirow{5}{*}{0.00}\\
      \cdashline{3-8}
    & & MRE-B-4 & What does sustainable energy mean?  & 0.00 & - & &\\
      \cdashline{3-8}
    & & MRE-R-L & What does it mean if energy is sustainable?  & - & 0.33 & &\\
      \cdashline{3-8}
    & & MRE-BS & What does sustainable energy mean?  & - & - & 0.76 & - \\
      \cdashline{3-8}
    & & MRE-BR & What does it mean if energy is sustainable?  & - & - & - & 0.77 \\
\bottomrule

\end{tabular}
}
\caption{Examples of SRE and MRE results. MRE-B-4, MRE-R-L, MRE-BS, and MRE-BR denotes to use BLEU-4, ROUGE-L, BERTScore, and BLEURT as \( M\), respectively. Reference Question for SRE represents the given reference question \( q^{r}_{0}\), and the Reference Question for MRE-B-4, MRE-R-L, MRE-BS, and MRE-BR represent one of \( \hat{Q}\) that obtained the max score for each measure.}
\vspace{-0.2cm}
\label{table:introduction}
\end{table*}

\subsection{Human Evaluation of Question Paraphrase}
\label{sec:human}

The assumption of multi-reference evaluation is that most paraphrased questions with LLMs can serve the meaning like the gold questions. We conduct a human study to validate this assumption.
For each of GPT-3 (0-shot), GPT-3 (3-shot), and ChatGPT, we sample 50 pairs of reference questions and paraphrased questions and annotate each pair whether the paraphrased questions have the same meaning and have the same answer compared to reference questions.
Specifically, we ask two annotators to evaluate with a binary rating (1 for "same" and 0 for "not same").
As shown in Table~\ref{tab:human-paraphrased}, 92\% and 93\% of the questions paraphrased by ChatGPT are evaluated as having the same answer and meaning, respectively.
In addition, even when paraphrasing with GPT3 3-shot, it has the same meaning and the same answer at a high rate.
We refer to Appendix~\ref{appsec:human} for more details about human annotation.

\subsection{Case Study}
\label{appsec:case-study}
For example in E1 in Table~\ref{table:introduction}, one of the texts in paraphrased references matches the generated question. 
MRE achieves gains over SRE by 1.00 ($0.00 \rightarrow 1.00$) on BLEU-4, and we found a positive effect on all other metrics.
In E2, the text that received the highest score among paraphrased references differs from each metric. We can observe that MRE works well by showing that you can choose one of the paraphrased references that are measured to be similar for each metric. Moreover, score increases suggest that MRE leads to positive shifts in the metric scores when the human score is 1 (E1, E2). However, the score to utilize MRE cannot be lower than SRE in any example because MRE takes the maximum score for the true reference and paraphrased references. Thus, if the human score is low, it is important to have a small negative effect.
One may ask about the risk of MRE giving a higher score than SRE for wrong questions as in E3. However, we argue that it doesn't weaken the strength of MRE as the gaps between SRE and MRE for wrong questions are relatively smaller than that for correct questions, which we compared in Table~\ref{table:ablation}.

\section{Conclusion \& Future Work}
\label{sec:conclusion}

In this paper, we studied the problem of evaluating the question generation frameworks, and observed that automatically augmenting the reference question with large language models is surprisingly effective, showing higher correlations with human-annotated scores. 
Though we evaluated the effectiveness of multiple reference evaluations for test-time evaluations, where the gold human score is given, we hope future research to explore other scenarios, such as measuring validation performance (asking how much the test performance can be actually improved) and multi-reference training as in~\cite{jeong2021label}. 
Exploring other tasks (machine translation and document summarization) or generation methods (giving context and the reference question together to LLMs) would be interesting for future research. 
\section{Limitations}
\label{sec:limitations}

\noindent\textbf{Inapplicability to reference-free evaluation}: Since our MRE supposes that there is an available reference question to be augmented (paraphrased), it is not applicable to reference-free question evaluations such as QRelScore~\cite{wang2022qrelscore} and RQUGE~\cite{mohammadshahi2022rquge}.

\noindent\textbf{Inapplicability for answer-unconditional QG frameworks}: MRE can't be applied to answer-unconditional QG frameworks because it only augments the reference question by paraphrasing without considering other possible questions of supposing other answers.

\noindent\textbf{Large computations}: 
To generate multi-reference questions, our method requires inference of large language models, which results in huge computational costs.
Therefore, this can become burdensome as the test dataset grows.

\section{Ethical Considerations}
\label{sec:ethical}
We honor and support the ACL code of Ethics.
In order to conduct our human annotation for paraphrased sentences, two humans are recruited.
We make sure that humans would be paid a wage of 15 dollars per hour.

\bibliography{custom}
\bibliographystyle{acl_natbib}

\appendix
\clearpage
\label{sec:appendix}
\section{LLM prompts}

\label{appsec:prompts}

In this section, we describe the details of how to paraphrase the questions with LLMs. To avoid confusion, we report it with an example question.

\label{appsec:summary-metrics}
\subsection{ChatGPT (0-shot) \& GPT-3 (0-shot)}
We use 0-shot input for ChatGPT and GPT-3 as:

\begin{small}
\begin{verbatim}
Please paraphrase the following sentence 20 times:
Where is the Statue of Liberty located?
\end{verbatim}
\end{small}

\subsection{GPT-3 (3-shot)}
For the 3-shot prompt, we use the examples paraphrased by ChatGPT as the 3-shot.
Our 3-shot prompt is described as:

\begin{small}
\begin{Verbatim}[breaklines=true]
Please paraphrase the following sentence 20 times:

Sentence: What is this software found useful for?
1. For what purpose is this software deemed useful?
2. What are the uses of this software?
3. Can you tell me what this software is useful for?
4. What are the benefits of this software?
5. In what ways is this software found to be useful?
6. What are the applications of this software?
7. Can you explain the usefulness of this software?
8. What does this software excel at?
9. Can you tell me what tasks this software is useful for?
10. In what scenario is this software useful?
11. Can you describe the utility of this software?
12. What is the purpose of this software?
13. For what is this software commonly used?
14. What does this software aid in?
15. Can you tell me the function of this software?
16. What are the advantages of this software?
17. What is this software good for?
18. In what ways does this software provide value?
19. What is the benefit of using this software?
20. Can you tell me what this software is commonly used for?

Sentence: Who is the girl?
1. Can you tell me the name of the young lady?
2. Who is the female in question?
3. Could you identify the girl for me?
4. To whom are you referring as the girl?
5. The girl, who is she?
6. Could you tell me who the girl is?
7. Who is the young woman being spoken of?
8. Can you name the girl in question?
9. Who is the lady in question?
10. Could you give me the name of the girl?
11. Who is being referred to as the girl?
12. Can you tell me who the female is?
13. Who is the girl being discussed?
14. Can you identify the young lady?
15. The girl, can you tell me her name?
16. Who is the subject of the girl?
17. Can you name the female in question?
18. Who is the girl you are asking about?
19. Can you provide the name of the girl?
20. Who is the young woman being referred to?

Sentence: Sentence: Where is the Eiffel Tower?
1. Can you tell me the location of the Eiffel Tower?
2. Could you inform me where the Eiffel Tower is situated?
3. I'm wondering where the Eiffel Tower is located?
4. The Eiffel Tower, where can I find it?
5. Could you give me the whereabouts of the Eiffel Tower?
6. The Eiffel Tower, where is it located?
7. Can you indicate the location of the Eiffel Tower?
8. Can you provide me with the location of the Eiffel Tower?
9. Where can I find the Eiffel Tower?
10. The Eiffel Tower, where is it situated?
11. Can you tell me where the Eiffel Tower is located?
12. Could you give me the location of the Eiffel Tower?
13. Where is the Eiffel Tower situated?
14. The Eiffel Tower, where is it found?
15. Could you inform me where the Eiffel Tower can be found?
16. Can you give me the whereabouts of the Eiffel Tower?
17. Where is the Eiffel Tower located?
18. The Eiffel Tower, where is it positioned?
19. Can you indicate the whereabouts of the Eiffel Tower?
20. Can you provide me with the whereabouts of the Eiffel Tower?

Sentence: 
\end{Verbatim}
\end{small}

\section{Human Annotation}
\label{appsec:human}
Two annotators participate in our study.
All the pairs from paraphrasing LLMs are randomly shuffled and anonymized, and each pair is evaluated by the following two dimensions:
\paragraph{Same Answer}
Human annotators check whether the paraphrased question has the same answer as the reference question.
Annotation is performed by binary rate, 1 for "having the same answer" and 0 for "having the different answer".
\paragraph{Same meaning}
It checks whether the paraphrased question has the same meaning as the reference question.
Humans annotate the question as 1 for "having the same meaning" and 0 for "having a different meaning". The inter-annotator agreement is 0.24 for the same meaning, and 0.21 for the same answer. Although the agreement was low due to the difference in their standards, the model preference was clearly preserved for both annotators.

 \end{document}